\documentclass[twoside,11pt]{article}
\pdfoutput=1

\usepackage{jmlr2e}

\usepackage{color}
\usepackage{fancyhdr}
\usepackage{microtype}
\usepackage{subfigure}
\usepackage{booktabs} 
\usepackage{enumitem}
\definecolor{Green}{RGB}{10,200,100}

\usepackage{hyperref}
\usepackage{amsmath}
\usepackage{amsfonts}

\usepackage{listings}
\usepackage[usenames,dvipsnames]{xcolor}

\definecolor{strings}{rgb}{.624,.251,.259}
\definecolor{keywords}{rgb}{.224,.451,.686}
\definecolor{comment}{rgb}{.322,.451,.322}

\lstdefinelanguage{python}{
  morekeywords={from, import, as, for, in, while, def, return, 
  =, +, -, /, *, lambda},
  keywords=[3]{sample, param, module, Marginal, Posterior, Trace, Poutine, Distribution},
  morecomment=[l]{\#},
  morecomment=[s]{"""}{"""},
  morestring=[b]',
  morestring=[b]",
  alsoletter={<>=-+/*},
  sensitive=true
}

\renewcommand{\texttt}[1]{\lstinline[basicstyle=\fontsize{8pt}{8.25pt}\selectfont\ttfamily]{#1}}

\jmlrheading{X}{XXXX}{X-X}{06/18}{XX/XX}{paper18a}{authors}  

\ShortHeadings{Pyro: Deep Universal Probabilistic Programming}{Bingham et al.}
\firstpageno{1}

\begin{document}

\title{Pyro: Deep Universal Probabilistic Programming}

\author{\name Eli Bingham \email eli.bingham@uber.com
       \AND
       \name Jonathan P. Chen \email jpchen@uber.com
       \AND
       \name Martin Jankowiak \email jankowiak@uber.com
       \AND
       \name Fritz Obermeyer \email fritzo@uber.com
       \AND
       \name Neeraj Pradhan \email npradhan@uber.com
       \AND
       \name Theofanis Karaletsos \email theofanis@uber.com
       \AND
       \name Rohit Singh \email rohits@uber.com
       \AND
       \name Paul Szerlip \email pas@uber.com \\
       \addr Uber AI Labs \\ Uber Technologies, Inc. \\ San Francisco, CA
       \AND
       \name Paul Horsfall \email horsfallp@gmail.com
       \AND
       \name Noah D. Goodman \email ngoodman@stanford.edu \\
       \addr Uber AI Labs \\ San Francisco, CA\\
       Stanford University, Stanford, CA, USA}

\editor{XXX}  

\maketitle

\begin{abstract}
  Pyro is a probabilistic programming language built on Python
  as a platform for developing advanced probabilistic models in AI research.
  To scale to large datasets and high-dimensional models,
  Pyro uses stochastic variational inference algorithms
  and probability distributions built on top of PyTorch,
  a modern GPU-accelerated deep learning framework.
  To accommodate complex or model-specific algorithmic behavior,
  Pyro leverages Poutine, a library of composable building blocks
  for modifying the behavior of probabilistic programs.
\end{abstract}


\newpage  

\begin{figure}[!htb]
\minipage{0.5\textwidth}
\begin{lstlisting}[language=python]
def model():
	loc, scale = torch.zeros(20), torch.ones(20)
	z = pyro.sample("z", Normal(loc, scale))
	w, b = pyro.param("weight"), pyro.param("bias")
	ps = torch.sigmoid(torch.mm(z, w) + b)
	return pyro.sample("x", Bernoulli(ps))


def guide(x):
  pyro.module("encoder", nn_encoder)
  loc, scale = nn_encoder(x)
  return pyro.sample("z", Normal(loc, scale))
\end{lstlisting}
\endminipage 
\hfill
\minipage{0.5\textwidth}
\begin{lstlisting}[language=python]
def conditioned_model(x):
	return pyro.condition(model, data={"x": x})()

optimizer = pyro.optim.Adam({"lr": 0.001})
loss = pyro.infer.Trace_ELBO()

svi = pyro.infer.SVI(model=conditioned_model, 
                     guide=guide, 
                     optim=optimizer, 
                     loss=loss)

losses = []
for batch in batches:
    losses.append(svi.step(batch))
\end{lstlisting}
\endminipage
\caption{A complete Pyro example: the generative model (\texttt{model}), approximate posterior (\texttt{guide}), constraint specification (\texttt{conditioned_model}), and stochastic variational inference (\texttt{svi}, \texttt{loss}) in a variational autoencoder.  \texttt{encoder} is a \texttt{torch.nn.Module} object. \texttt{pyro.module} calls \texttt{pyro.param} on every parameter of a \texttt{torch.nn.Module}.}
\label{vae_example}
\end{figure}
\section{Introduction}

In recent years, richly structured probabilistic models have demonstrated promising results on a number of fundamental problems in AI
(
\cite{ghahramani_probabilistic_2015}).
However, most such models are still implemented from scratch as one-off systems,
slowing their development and limiting their scope and extensibility.
Probabilistic programming languages (PPLs) promise to reduce this burden, but
in practice more advanced models often require high-performance
inference engines tailored to a specific application.
We identify design principles that enable a PPL
to scale to advanced research applications while retaining flexibility,
and we argue that these principles are fully realized together in the Pyro PPL.

\section{Design Principles}
First, a PPL suitable for developing state-of-the-art AI research models should be \textbf{expressive}:
it should be able to concisely describe models with have data-dependent internal control flow or latent variables whose existence depends on the values of other latent variables, or models which only be defined in closed form as unnormalized joint distributions.

For a PPL to be practical, it must be \textbf{scalable}:
its approximate inference algorithms
must be able to seamlessly handle the large datasets and non-conjugate, high-dimensional models common in AI research,
and should exploit 
compiler acceleration when possible.

A PPL targeting research models should be \textbf{flexible}:
in addition to scalability, many advanced models require inference algorithms
with complex, model-specific behavior.
A PPL should enable researchers to quickly and easily implement such behavior
and should enforce a separation of concerns between 
between model, inference, and runtime implementations.

Finally, a PPL targeting researchers as users should strive to be \textbf{minimal}:
in order to minimize cognitive overhead,
it should share most of its syntax and semantics with existing languages and 
and systems and work well with other tools such as libraries for visualization.

As is clear from Table \ref{ppl_comparison}, these four principles are often in conflict, with one being achieved at the expense of others. 
For example, an overly flexible design may be very difficult to implement efficiently and scalably,
especially while simultaneously integrating a new language with existing tools. Similarly, enabling development of custom
inference algorithms may be difficult without limiting model expressivity.
In this section, we describe the design choices we made in Pyro to balance between all four objectives.

Pyro is embedded in Python,
and Pyro programs are written as Python functions, or callables, with just two extra language primitives (whose behavior is overridden by inference algorithms): \texttt{pyro.sample} for annotating calls to functions with internal randomness, and \texttt{pyro.param} for registering learnable parameters with inference algorithms that can change them.
Pyro models may contain arbitrary Python code
and interact with it in arbitrary ways, including expressing unnormalized models through the \texttt{obs} keyword argument to \texttt{pyro.sample}.
Pyro's language primitives may be used with all of Python's
control flow constructs, including recursion, loops, and conditionals. The existence of random variables in a particular execution
may thus depend on any Python control flow construct.

Pyro implements several generic probabilistic inference algorithms,
including the No U-turn Sampler (\cite{hoffman_no-u-turn_2014}), a variant of Hamiltonian Monte Carlo. 
However, the primary inference algorithm is gradient-based stochastic variational inference (SVI) 
(\cite{kingma2013auto}),
which uses stochastic gradient descent to optimize Monte Carlo estimates of 
a divergence measure between approximate and true posterior distributions. 
Pyro scales to complex, high-dimensional models thanks to GPU-accelerated tensor math and reverse-mode automatic differentiation via PyTorch,
and it scales to large datasets thanks to 
stochastic gradient 
estimates computed over mini-batches of data in SVI.

Some inference algorithms in Pyro, such as SVI and importance sampling,
can use arbitrary Pyro programs 
(called \emph{guides}, following webPPL)
as approximate posteriors or proposal distributions. 
A guide for a given model must take the same input arguments as the model 
and contain a corresponding sample statement for every unconstrained
sample statement in the model but is otherwise unrestricted.
Users are then free to express complex hypotheses
about the posterior distribution, e.g.~its conditional independence structure.
Unlike webPPL and Anglican, in Pyro guides may not depend on values inside the model.

Finally, to achieve flexibility and separation of concerns,
Pyro is built on Poutine, a library of effect handlers (\cite{kammar_handlers_2013})
that implement individual control and book-keeping operations
used for inspecting and modifying the behavior of Pyro programs,
separating inference algorithm implementations from language details.

\section{Project Openness and Development}  

Pyro's source code is freely available under an MIT license and developed by the authors and a community of open-source contributors  
at \url{https://github.com/uber/pyro} and documentation, examples, and a discussion forum
are hosted online at \url{https://pyro.ai}.
A comprehensive test suite is run automatically by a continuous integration
service before code is merged into the main codebase to maintain a high level of project quality and usability.

We also found that while PyTorch was an 
invaluable substrate for tensor operations and automatic differentiation,
it was lacking a high-performance library of probability distributions.
As a result, several of the authors made substantial open-source
contributions upstream to PyTorch Distributions,\footnote{\url{http://pytorch.org/docs/stable/distributions.html}} 
a new PyTorch core library inspired by TensorFlow Distributions (\cite{dillon_tensorflow_2017}).
\section{Existing Systems}
\begin{figure}[!htb]
\begin{center}
\begin{tabular}{l*{4}{c}r}
System & \parbox[b]{3cm}{Expressivity:\\Dynamic control} & \parbox[b]{3cm}{Scalability:\\Subsampling, AD} & \parbox[b]{3cm}{Flexibility:\\Flexible inference} & \parbox[b]{3cm}{Minimality:\\Host language} \\
\hline
Stan  & Static control flow  & Some, CPU & Automated & None\\
Church  & Yes & No, None  & Automated  & Scheme \\
Venture & Yes & No, None & Yes  & None \\
webPPL  & Yes & No, CPU & Some & JavaScript \\
Edward  & Static control flow & Yes, CPU/GPU & Yes & TensorFlow \\
Pyro    & Yes & Yes, CPU/GPU & Yes & Python, PyTorch \\
\end{tabular}
\end{center}
\caption{Simplified summary of design principles of Pyro and some other PPLs.}
\label{ppl_comparison}
\end{figure}
Probabilistic programming and approximate inference are areas of active research,
so there are many existing probabilistic programming languages and systems.
We briefly mention several that were especially influential in Pyro's development,
regretfully omitting (due to space limitations) many systems for which simple direct comparisons are more difficult. We also emphasize that, as in conventional programming language development,
our design decisions are not universally applicable or desirable
for probabilistic programming, and that other systems purposefully
make different tradeoffs to achieve different goals.

Stan (\cite{carpenter_stan_2017}) is a domain-specific language designed for describing
a restricted class of probabilistic programs and performing high-quality automated inference in those models.
Church (\cite{goodman_church:_2008}), a probabilistic dialect of Scheme,
was an early universal probabilistic programming language,
capable of representing any computable probability distribution.
Venture (\cite{mansinghka_venture:_2014}) is a universal language with a focus on expressiveness and flexibility and a custom syntax and virtual machine.
Anglican (\cite{tolpin_design_2016})
and webPPL (\cite{goodman_design_2014})
are lightweight successors to Church,
embedded as syntactic subsets (\cite{wingate_lightweight_2011})
in the general-purpose programming languages
Clojure and JavaScript.  
Edward (\cite{tran_deep_2017}) is a PPL built on static
TensorFlow graphs that features composable representations
for models and inference.  
ProbTorch (\cite{siddharth_learning_2017}) is a PPL built on PyTorch with a focus on deep generative models and developing new objectives for variational inference.
Turing (\cite{ge_turing_2018}) is a PPL embedded in Julia featuring composable MCMC algorithms.

\section{Experiments}
\begin{figure}[!htb]
\minipage{0.4\textwidth}
\begin{center}
\begin{tabular}{| c | c | c | c |}
\hline
\# $z$ & \# $h$ & PyTorch (ms) & Pyro (ms) \\
\hline
10 & 400 & $3.82 \pm 0.02$ & $6.79 \pm 0.04$ \\ 
30 & 400 & $3.73 \pm 0.07$ & $6.67 \pm 0.03$ \\
10 & 2000 & $7.65 \pm 0.02$ & $10.14 \pm 0.06$ \\ 
30 & 2000 & $7.66 \pm 0.02$ & $10.19 \pm 0.03$ \\ 
\hline
\end{tabular}
\end{center}
\caption{Times per update of VAE in Pyro versus PyTorch}
\label{vae_experiment}
\endminipage
\hspace{0.10\textwidth}
\minipage{0.4\textwidth}
\begin{center}
\begin{tabular}{| l | c |}
\hline
\# IAFs & Test ELBO \\
\hline
0 (theirs) & -6.93 \\
\hline
0 (ours) & -6.87 \\
1 & -6.82 \\
2 & -6.80 \\
\hline
\end{tabular}
\end{center}
\caption{Test ELBOs for DMM and extension with IAF guide}
\label{dmm_experiment}
\endminipage
\end{figure}

To demonstrate that Pyro meets our design goals,
we implemented several state-of-the-art models.\footnote{\url{http://pyro.ai/examples/}}
Here we focus on the variational autoencoder (VAE; \cite{kingma2013auto}) and the Deep Markov Model (DMM; \cite{krishnan_structured_2017}),
a non-linear state space model that has been used
for several applications including audio generation and causal inference.
The VAE is a standard example in deep probabilistic modeling, while the DMM has several characteristics that make it ideal as a point of comparison: it is a high-dimensional, non-conjugate model designed to be fit to large datasets; the number of latent variables in a sequence depends on the input data; and it uses a hand-designed approximate posterior.

The models and inference procedures derived by Pyro replicate the original papers almost exactly, except that we use Monte Carlo estimates rather than exact analytic expressions for KL divergence terms.
We use the MNIST dataset and 2-hidden-layer MLP encoder and decoder networks with varying hidden layer size $\# h$ and latent code size $\# z$ for the VAE
and the same dataset of digitized music\footnote{This is the JSB chorales dataset from \texttt{http://www-etud.iro.umontreal.ca/\~Eboulanni/icml2012}}
to train the DMM.

To demonstrate that Pyro's abstractions do not reduce its scalability by introducing too much overhead,
we compared our VAE implementation with an idiomatic PyTorch
implementation.\footnote{We used the PyTorch example at \url{https://github.com/pytorch/examples/tree/master/vae}}  After verifying that they converge to the same test ELBO, 
we compared the wall-clock time
taken to compute one gradient update, averaged over 10 epochs of GPU-accelerated mini-batch stochastic
gradient variational inference (batch size 128) on a single NVIDIA GTX 1080Ti.  Figure \ref{vae_experiment} shows that
the relative performance gap between the Pyro and PyTorch versions is moderate, and, more importantly, that it shrinks as the total time spent performing tensor operations increases.

We used the DMM to evaluate Pyro's flexibility and expressiveness. As Figure \ref{dmm_experiment} shows, 
we found that we were able to quickly and concisely replicate the exact 
DMM model and inference configuration and 
quantitative results reported in the paper after 5000 training epochs.
Furthermore, thanks to Pyro's modular design, 
we were also able to build DMM variants
with more expressive approximate posteriors via autoregressive flows (IAFs) (\cite{kingma_improved_2016}),
improving the results with a few lines of code
at negligible computational cost.

\acks{We would like to acknowledge Du Phan, Adam Scibior, Dustin Tran, Adam Paszke, Soumith Chintala, Robert Hawkins, Andreas Stuhlmueller, our colleagues in Uber AI Labs, and the Pyro and PyTorch open-source communities for helpful contributions and feedback.}

\bibliography{pyro_bib2}

\begin{thebibliography}{16}
\providecommand{\natexlab}[1]{#1}
\providecommand{\url}[1]{\texttt{#1}}
\expandafter\ifx\csname urlstyle\endcsname\relax
  \providecommand{\doi}[1]{doi: #1}\else
  \providecommand{\doi}{doi: \begingroup \urlstyle{rm}\Url}\fi

\bibitem[Carpenter et~al.(2017)Carpenter, Gelman, Hoffman, Lee, Goodrich,
  Betancourt, Brubaker, Guo, Li, and Riddell]{carpenter_stan_2017}
Bob Carpenter, Andrew Gelman, Matthew~D. Hoffman, Daniel Lee, Ben Goodrich,
  Michael Betancourt, Marcus Brubaker, Jiqiang Guo, Peter Li, and Allen
  Riddell.
\newblock Stan: {A} {Probabilistic} {Programming} {Language}.
\newblock \emph{Journal of Statistical Software}, 76\penalty0 (1), 2017.

\bibitem[Dillon et~al.(2017)Dillon, Langmore, Tran, Brevdo, Vasudevan, Moore,
  Patton, Alemi, Hoffman, and Saurous]{dillon_tensorflow_2017}
Joshua~V. Dillon, Ian Langmore, Dustin Tran, Eugene Brevdo, Srinivas Vasudevan,
  Dave Moore, Brian Patton, Alex Alemi, Matt Hoffman, and Rif~A. Saurous.
\newblock {TensorFlow} {Distributions}.
\newblock \emph{arXiv:1711.10604}, November 2017.

\bibitem[Ge et~al.(2018)Ge, Xu, and Ghahramani]{ge_turing_2018}
Hong Ge, Kai Xu, and Zoubin Ghahramani.
\newblock {Turing: A Language for Flexible Probabilistic Inference}.
\newblock In \emph{AISTATS}, 2018.

\bibitem[Ghahramani(2015)]{ghahramani_probabilistic_2015}
Zoubin Ghahramani.
\newblock Probabilistic machine learning and artificial intelligence.
\newblock \emph{Nature}, 521:\penalty0 452--459, May 2015.

\bibitem[Goodman and Stuhlm\"{u}ller(2014)]{goodman_design_2014}
Noah~D Goodman and Andreas Stuhlm\"{u}ller.
\newblock {The Design and Implementation of Probabilistic Programming
  Languages}.
\newblock \url{http://dippl.org}, 2014.

\bibitem[Goodman et~al.(2008)Goodman, Mansinghka, Roy, Bonawitz, and
  Tenenbaum]{goodman_church:_2008}
Noah~D. Goodman, Vikash~K. Mansinghka, Daniel Roy, Keith Bonawitz, and
  Joshua~B. Tenenbaum.
\newblock Church: {A} {Language} for {Generative} {Models}.
\newblock In \emph{UAI}, 2008.

\bibitem[Hoffman and Gelman(2014)]{hoffman_no-u-turn_2014}
Matthew~D. Hoffman and Andrew Gelman.
\newblock The {No}-{U}-turn {Sampler}: {Adaptively} {Setting} {Path} {Lengths}
  in {Hamiltonian} {Monte} {Carlo}.
\newblock \emph{J. Mach. Learn. Res.}, 15\penalty0 (1), January 2014.

\bibitem[Kammar et~al.(2013)Kammar, Lindley, and Oury]{kammar_handlers_2013}
Ohad Kammar, Sam Lindley, and Nicolas Oury.
\newblock Handlers in {Action}.
\newblock In \emph{ICFP}, 2013.

\bibitem[Kingma and Welling(2014)]{kingma2013auto}
Diederik~P Kingma and Max Welling.
\newblock Auto-encoding {Variational} {Bayes}.
\newblock In \emph{ICLR}, 2014.

\bibitem[Kingma et~al.(2016)Kingma, Salimans, Jozefowicz, Chen, Sutskever, and
  Welling]{kingma_improved_2016}
Diederik~P Kingma, Tim Salimans, Rafal Jozefowicz, Xi~Chen, Ilya Sutskever, and
  Max Welling.
\newblock Improved {Variational} {Inference} with {Inverse} {Autoregressive}
  {Flow}.
\newblock In \emph{NIPS}. 2016.

\bibitem[Krishnan et~al.(2017)Krishnan, Shalit, and
  Sontag]{krishnan_structured_2017}
Rahul~G Krishnan, Uri Shalit, and David Sontag.
\newblock Structured {Inference} {Networks} for {Nonlinear} {State} {Space}
  {Models}.
\newblock In \emph{{AAAI}}, 2017.

\bibitem[Mansinghka et~al.(2018)Mansinghka, Schaechtle, Handa, Radul, Chen, and
  Rinard]{mansinghka_venture:_2014}
Vikash~K. Mansinghka, Ulrich Schaechtle, Shivam Handa, Alexey Radul, Yutian
  Chen, and Martin Rinard.
\newblock Probabilistic {Programming} with {Programmable} {Inference}.
\newblock In \emph{PLDI}, 2018.

\bibitem[Siddharth et~al.(2017)Siddharth, Paige, van~de Meent, Desmaison,
  Goodman, Kohli, Wood, and Torr]{siddharth_learning_2017}
N.~Siddharth, Brooks Paige, Jan-Willem van~de Meent, Alban Desmaison, Noah~D.
  Goodman, Pushmeet Kohli, Frank Wood, and Philip Torr.
\newblock {Learning Disentangled Representations with Semi-Supervised Deep
  Generative Models}.
\newblock In \emph{NIPS}, 2017.

\bibitem[Tolpin et~al.(2016)Tolpin, van~de Meent, Yang, and
  Wood]{tolpin_design_2016}
David Tolpin, Jan-Willem van~de Meent, Hongseok Yang, and Frank Wood.
\newblock {Design and Implementation of Probabilistic Programming Language
  Anglican}.
\newblock In \emph{IFL}, 2016.

\bibitem[Tran et~al.(2017)Tran, Hoffman, Saurous, Brevdo, Murphy, and
  Blei]{tran_deep_2017}
Dustin Tran, Matthew~D. Hoffman, Rif~A. Saurous, Eugene Brevdo, Kevin Murphy,
  and David~M. Blei.
\newblock {Deep Probabilistic Programming}.
\newblock In \emph{ICLR}, 2017.

\bibitem[Wingate et~al.(2011)Wingate, Stuhlmüller, and
  Goodman]{wingate_lightweight_2011}
David Wingate, Andreas Stuhlmüller, and Noah Goodman.
\newblock {Lightweight Implementations of Probabilistic Programming Languages
  via Transformational Compilation}.
\newblock In \emph{AISTATS}, 2011.

\end{thebibliography}

\end{document}